\documentclass[sigconf]{acmart}
\renewcommand\footnotetextcopyrightpermission[1]{}
\AtBeginDocument{%
  }

\setcopyright{acmlicensed}
\copyrightyear{2025}
\acmYear{2025}
\acmDOI{XXXXXXX.XXXXXXX}
\acmConference[ACM Conference]{Make sure to enter the correct conference title from your rights confirmation email}{2026}{}

\acmSubmissionID{2067}

\usepackage{subcaption}
\usepackage{cleveref}

\usepackage[table,xcdraw]{xcolor}

\crefname{figure}{Fig.}{Figs.}
\crefname{table}{Tab.}{Tabs.}
\crefname{section}{Sec.}{Secs.}
\crefname{subsection}{Sec.}{Secs.}
\crefname{equation}{Eq.}{Eqs.}
\crefname{appendix}{Appendix}{Appendices}

\usepackage{titlesec}
\titleformat{\subsubsection}
  [hang]                
  {\large\bfseries}     
  {}                    
  {0em}                 
  {\noindent}           
  []                    

\titlespacing*{\subsubsection}
  {0em}    
  {0.5em}  
  {0.3em}  

\begin{document}

\title{Learning Physics from Pretrained Video Models: A Multimodal Continuous and Sequential World Interaction Models for Robotic Manipulation}


\author{
  Zijian Song\textsuperscript{1},
  Qichang Li\textsuperscript{1},
  Sihan Qin\textsuperscript{1},
  Yuhao Chen\textsuperscript{1},\\
  Tianshui Chen\textsuperscript{3,4},
  Liang Lin\textsuperscript{1,2,3},
  Guangrun Wang\textsuperscript{1,2,3,*}
  \\
  \small{
    \textbf{Email:} \{songzj8, liqch33, qinsh9, chenyh387\} (at) mail2.sysu.edu.cn, chentianshui (at) gdut.edu.cn, linliang (at) ieee.org, wanggrun (at) gmail.com
  }
}

\affiliation{
    \institution{
        \textsuperscript{1}Sun Yat-sen University; \textsuperscript{2}Guangdong Key Laboratory of Big Data Analysis and Processing;\\
        \textsuperscript{3}X-Era AI Lab;\textsuperscript{4}Guangdong University of Technology
    }
    \city{}
    \country{}
}

\renewcommand{\shortauthors}{}

\begin{abstract}
The scarcity of large-scale robotic data has motivated the repurposing of foundation models from other modalities for policy learning. In this work, we introduce PhysGen (Learning Physics from Pretrained Video Generation Models), a scalable continuous and sequential world interaction framework that leverages autoregressive video generation to solve robotic manipulation tasks. By treating the pretrained video model as a proxy for a physics simulator, PhysGen models the dynamic interplay between the external environment and robot actions. We introduce a multimodal continuous representation that unifies video and action into shared physical tokens, bridging the gap between discrete video generation and continuous robotic control. This approach enables the seamless transfer of implicit physical knowledge—such as object permanence and dynamics—from video pretraining to downstream manipulation.To ensure efficient convergence, we incorporate causal masking, inverse kinematics, Lookahead Multi-Token Prediction (L-MTP), and key-value (KV) caching. Experimental results on the Libero and ManiSkill benchmarks demonstrate that PhysGen consistently outperforms robust baselines, surpassing OpenVLA and WorldVLA by margins of 13.8\% and 8.8\%, respectively. Notably, in real-world scenarios, PhysGen matches the performance of large-scale action-pretrained models like $\pi_0$ without requiring prior action-specific pretraining, demonstrating superior capability in physically complex tasks such as grasping transparent objects. These findings validate the potential of extracting physical intuition from pretrained video generators to facilitate generalizable robotic manipulation.

\end{abstract}




\maketitle

\begingroup
  \renewcommand\thefootnote{\fnsymbol{footnote}} 
  \footnotetext[1]{Corresponding author: Guangrun Wang}
\endgroup

\renewcommand{\floatpagefraction}{0.8}
\section{Introduction}
\label{sec:intro}

The remarkable success of large-scale generative pretraining in computer vision and natural language processing has demonstrated strong cross-task generalization~\cite{brown2020language, kirillov2023segment, tian2024visual, yuan2024sd, guo2025deepseek,xu2026bridging,chen2025style4d,song2025human,li2025situ,chen2025uml}. This has motivated the robotics community to seek similar foundation models for visuomotor control and manipulation. Yet, generative action pretraining faces unique challenges, as acquiring large-scale human demonstration data is both time-consuming and labor-intensive~\cite{zitkovich2023rt}. Consequently, a promising direction is to transfer knowledge from existing pretrained models to the action domain~\cite{zitkovich2023rt}. Recent works have predominantly built action policies upon large language models (LLMs), giving rise to Vision-Language-Action Models (VLAs)~\cite{zitkovich2023rt, kim2024openvla, black2024pi0, bu2025agibot, wang2025vla, zhan2025mathcal, li2025vla, chen2025villa, zhan2026stable}.

\begin{figure}[t]
    \centering
    \includegraphics[width=1.0\linewidth]{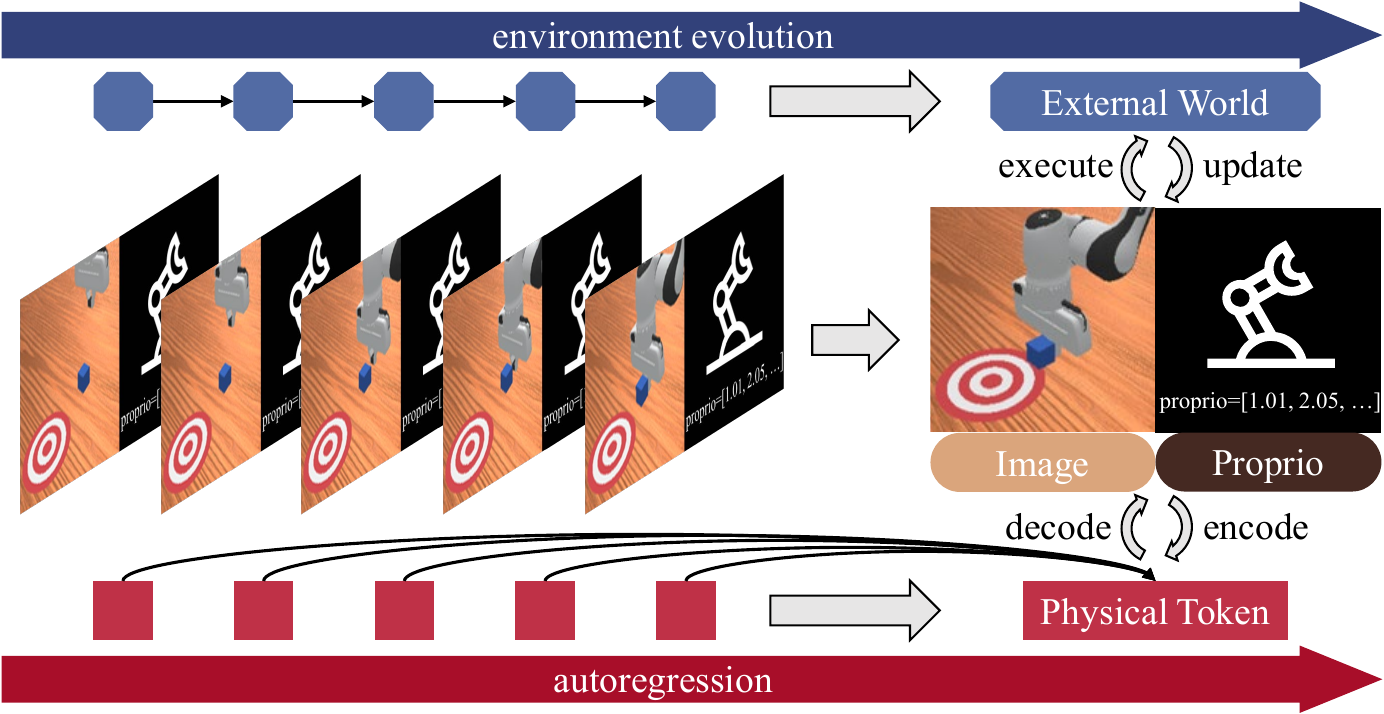}
    \caption{\underline{The PhysGen Framework}: Repurposing Video Generation as a World Simulator.
Our approach unifies perception and control through \textit{Physical Autoregression}. The model operates on a sequence of continuous \textbf{physical tokens} (red), which fuse visual context (orange) and embodiment actions (black) to predict their joint evolution.
Acting as a predictive world model, PhysGen runs in synchronization with the physical environment (blue): at each step, predicted tokens are decoded into executed actions and anticipated video frames. The resulting environmental feedback is re-encoded into the stream, enabling a seamless continuous loop of planning and interaction.}
    \label{fig:world_evolve}
\end{figure}

However, the intrinsic gap between text and action modalities often leads to suboptimal alignment between symbolic reasoning and physical control~\cite{luo2025beingh0}. Text describes the world symbolically, whereas manipulation requires precise spatial and temporal understanding. A more principled alternative is to build upon pretrained video generation models, particularly autoregressive ones~\cite{gao2024vid, xie2025progressive}. These models iteratively predict future observations from past context, implicitly capturing the physical laws and temporal dynamics of the world. Such models naturally serve as predictive world models; their step-by-step state evolution closely mirrors the sequential decision-making process in control, enabling the efficient transfer of learned world knowledge to action generation.

Based on these insights, we propose \textbf{PhysGen} (Learning Physics from Pretrained Video Generation Models), a unified world–action autoregressive framework that repurposes pretrained video generators as predictive physics simulators for robotic control. As illustrated in~\cref{fig:world_evolve}, the design of PhysGen is underpinned by two key pillars. First, it utilizes a pretrained autoregressive video backbone to inherit rich physical and temporal knowledge learned from large-scale video data, significantly improving data efficiency for downstream manipulation. Second, it introduces physical tokens that jointly encode video observations and actions. This enables the co-evolution of perception and action within a single autoregressive process, explicitly modeling the coupled dynamics between the robot and its environment.

Beyond the formulation of the framework, token representation poses a critical challenge~\cite{yu2023language, vuong2025action}. Conventional autoregressive models rely on discrete tokenization~\cite{zitkovich2023rt, kim2024openvla, wang2025unified}, which introduces resolution errors for continuous signals such as images and actions—errors that can accumulate over time and lead to trajectory drift~\cite{zhang2025chain}. While recent works have explored continuous representations~\cite{li2024autoregressive, deng2024autoregressive}, their integration into a unified vision-action framework remains limited. To bridge this gap, PhysGen represents both actions and visual observations as continuous tokens, modeled via a diffusion process to estimate their unconstrained probability density. This design enables a unified perception–planning–execution loop within a shared continuous embedding space. Unlike deterministic regression methods~\cite{wu2023unleashing, fu2024context, zhang2025chain}, PhysGen retains the generative expressiveness of autoregressive modeling, effectively capturing multimodal action distributions.

To further refine our framework, we incorporate three architectural designs:
(1) Causal Masking for Implicit Inverse Kinematics: We introduce a causal mask scheme for joint image–action autoregression, allowing action tokens to attend to future visual states, thereby fostering implicit inverse-kinematics reasoning.
(2) Lookahead-MTP: We combine lookahead formulation~\cite{chi2023diffusion, black2024pi0} with Multi-Token Prediction (MTP)\cite{gloeckle2024better, liu2024deepseek}, a strategy we term Lookahead-MTP (L-MTP). By decoding multiple future tokens in parallel, L-MTP extends the planning horizon and enhances temporal coherence.
(3) Efficient Training \& Inference: We adopt parallelized training with LoRA fine-tuning to preserve pretrained capabilities and employ a KV-cache mechanism\cite{NVIDIA2023LLM} for efficient real-time inference.

Simulation and real-world experiments demonstrate the efficacy of PhysGen. In simulation, we evaluate our method on the widely used LIBERO~\cite{liu2023libero} and ManiSkill~\cite{taomaniskill3} benchmarks. PhysGen consistently outperforms robust baselines, surpassing OpenVLA~\cite{kim2024openvla} and WorldVLA~\cite{cen2025worldvla} by margins of 13.8\% and 8.8\%, respectively. We further validate PhysGen on real-world manipulation tasks, where it matches or exceeds representative prior approaches—including large-scale action-pretrained models like Pi0\cite{black2024pi0}—without requiring prior action-specific pretraining. Qualitative visualizations show that PhysGen accurately predicts future video frames with tightly aligned action trajectories, particularly in physically complex tasks such as grasping transparent objects.

Our contributions are summarized as follows:
\begin{enumerate}
\item We introduce PhysGen, a physical autoregressive framework that repurposes video generation models for robotic control. By combining frames and actions into shared physical tokens, we inherit world knowledge from video pretraining to capture iterative physical dynamics, enabling physically grounded and robust manipulation.
\item We propose a continuous tokenization scheme where frames and actions are represented as continuous vectors. Their distributions are modeled through a DiT architecture and diffusion loss, enabling fine-grained control and deep interaction between continuous action and image generation.
\item We design a suite of architectural enhancements—including causal masking, Lookahead-MTP, and efficient KV-caching. Experiments on ManiSkill and LIBERO validate that PhysGen establishes a new state-of-the-art for video-based policy learning.
\end{enumerate}
\section{Related Work}
\label{sec:related_work}




\begin{figure*}[t]
    \centering
    \begin{minipage}[b]{0.8\linewidth}
        \centering
        \includegraphics[width=1.0\linewidth]{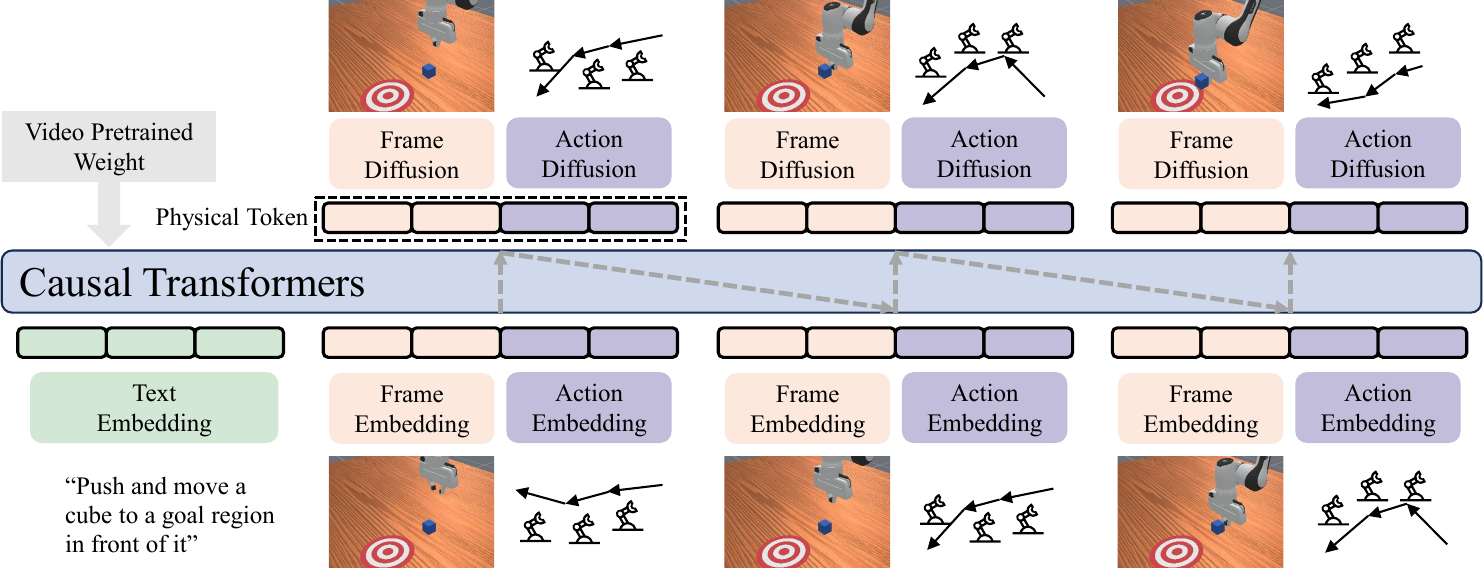}
    \end{minipage}
    \hfill
    \begin{minipage}[b]{0.18\linewidth}
        \centering
        \includegraphics[width=1.0\linewidth]{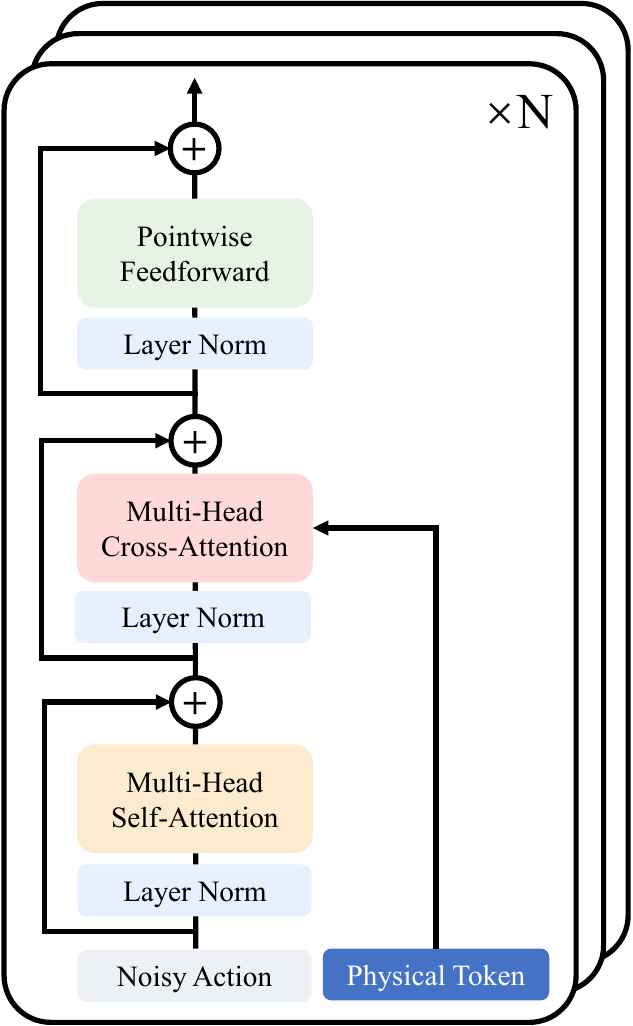}
    \end{minipage}
    \caption{\underline{The model architecture of PhysGen.} Starting from text tokens, PhysGen leverages a Causal Transformer to autoregressively predict physical tokens. Notably, we apply a frame diffusion and an action diffusion to estimate the conditional distributions of visual and action signals in continuous space.
    The structure of the action diffusion network is shown on the right column, where the predicted token serves as the conditional vector and is injected through cross-attention.}
    \label{fig:ar_transformer}
\end{figure*}

\subsection{Vision-Language-Action Models}
The scarcity of manipulation demonstrations makes large-scale action pretraining prohibitively expensive. A recent trend in robotics involves transferring pretrained knowledge from LLMs to the action domain~\cite{zitkovich2023rt, kim2024openvla, black2024pi0, zhan2025mathcal, zhang2025reasoning, bjorck2025gr00t, yuan2025autodrive, li2025vla, song2026robotic}. 
These approaches typically append a lightweight action head to the LLM, mapping language tokens to action outputs~\cite{wen2025tinyvla, wen2025dexvla, liu2025hybridvla, bu2025univla, wang2025vla}.
Motivated by this paradigm, we similarly adopt an autoregressive architecture, but build our model upon a pretrained video generation backbone rather than an LLM.
This design enables a seamless transfer of physical knowledge that aligns actions with the world dynamics captured by the video.

\subsection{Video-Action Joint Prediction}
The concept of jointly predicting video and actions has been explored in prior studies~\cite{wu2023unleashing, cheang2024gr, ni2024generate, cen2025worldvla, yelatent, li2025unified, liang2025video, shen2025videovla}.
For instance, WorldVLA~\cite{cen2025worldvla} unifies both image and action understanding and
generation. UWM~\cite{zhu2025unified} integrates an action diffusion process and a
video diffusion process within a unified transformer. UniMimic~\cite{chen2025unifying} jointly pretrains latent actions and latent states from videos.
Building on this line of research, our work employs an autoregressive framework that models the coupled evolution of the world and the robot in a step-by-step manner.
Uniquely, our approach departs from conventional approaches by representing frames and actions with a shared continuous token representation.
This unified representation facilitates a more direct transfer of pretrained world knowledge, effectively bridging the gap between visual dynamics and robotic control.

\subsection{Non-Quantized Autoregressive Models}
For continuous signals such as images or actions, quantization may disrupt their inherent structure and introduce resolution errors~\cite{zhang2025chain, kim2025fine}.
To preserve continuity for action generation, some approaches employ linear or MLP projectors but fall short for generative modeling~\cite{wu2023unleashing, fu2024context, zhang2025chain}.
More fundamentally, Li et al.~\cite{li2024autoregressive} proposed using a denoising process to compute the conditional distribution of signals, agnostic to the signal distribution. This idea has been extended to video generation~\cite{deng2024autoregressive} and action prediction~\cite{pi05vision, li2025unified, team2024octo, zhong2025dexgraspvla}.
However, its integration into a unified vision-action autoregressive framework remains underexplored. In this work, we leverage a denoising process as a de-tokenizer for continuous visual-action autoregression, fostering interactions between two modalities.
\section{Preliminaries}
\label{sec:preliminaries}


\subsection{Diffusion Models}  
Diffusion Models\cite{ho2020denoising} are a class of generative models that produce samples by progressively adding Gaussian noise to clean data \(x_0\) and then learning to invert this noising process. Formally, the forward diffusion process is defined as:
\begin{equation}
q(x_t \mid x_{t-1})
= \mathcal{N}\bigl(x_t;\sqrt{1-\beta_t}\,x_{t-1},\,\beta_t I\bigr),
\end{equation}
where \(x_t\) is the noisy sample at timestep \(t\), \(\beta_t\) is the pre-defined noise schedule, and \(\epsilon\sim\mathcal{N}(0,I)\) is the injected noise. The reverse diffusion process is defined as:
\begin{equation}
p_\theta(x_{t-1} \mid x_t)
= \mathcal{N}\bigl(x_{t-1};\mu_\theta(x_t,t),\,\Sigma_\theta(x_t,t)\bigr),
\end{equation}
which inverts the corruption, starting from the standard normal prior \(x_T\sim\mathcal{N}(0,I)\) at the final timestep \(T\). Training then teaches the model to remove noise by minimizing
\begin{equation}
L(\theta) = \mathbb{E}_{t,x_{0},\epsilon}\left[ \left\| \epsilon - \epsilon_{\theta}(\sqrt{\overline{\alpha}_{t}}x_{0} + \sqrt{1 - \overline{\alpha}_{t}}\epsilon, t) \right\|^{2} \right],
\end{equation}
where $\epsilon_{\theta}$ is the noise prediction from network $\theta$.

\subsection{Autoregression without Quantization}
Li et al.~\cite{li2024autoregressive} introduce a continuous autoregressive framework that directly models continuous signal distributions using diffusion loss, eliminating the need for quantization or discretization.
Concretely, given a ground-truth token embedding \(x\in\mathbb{R}^{d}\) and its autoregressive context \(z\), a lightweight denoiser \(\epsilon_{\theta}(x_{t}|t,z)\) is trained to predict the injected noise by minimizing the squared-error loss:
\begin{equation}
\mathcal{L}(z,x)=\mathbb{E}_{\epsilon,t}[||\epsilon-\epsilon_{\theta}(x_{t}|t,z)||^{2}].
\label{eq:diffusion_loss}
\end{equation}
Here, \(\epsilon\in\mathbb{R}^{d}\) is a noise vector sampled from \(\mathcal{N}(0,I)\). The corrupted vector is \(x_{t} = \sqrt{\overline{\alpha}_{t}}x+\sqrt{1-\overline{\alpha}_{t}}\epsilon\), where \(\overline{\alpha}_{t}\) defines a noise schedule and \(t\) is time step. \(\epsilon_{\theta}(x_{t}|t,z)\) means that the denoising network takes the corrupted token \(x_t\) as input and is conditioned on time step \(t\) and context \(z\).

At inference, each token is sampled by running the reverse chain, starting with \(x_{T}\sim\mathcal{N}(0,I)\):
\begin{equation}
x_{t-1}=\frac{1}{\sqrt{\alpha_{t}}}\left(x_{t}-\frac{1-\alpha_{t}}{\sqrt{1-\overline{\alpha}_{t}}}\epsilon_{\theta}(x_{t}|t,z)\right)+\sigma_{t}\delta.
\end{equation}
This reverse diffusion process ultimately produces a final sample \(x_{0}\sim p(x|z)\), representing a high-quality continuous embedding.

\subsection{Video Autoregressive Model}
NOVA~\cite{deng2024autoregressive} extends the above principle to model video generation in a continuous, non-quantized space, serving as the pretrained foundation for our method.

First, the temporal autoregressive process can be formally expressed by the following conditional probability, where $l$ denotes the prompt, \(S_{n}\) is the n-th frame token.:
\begin{equation}
p(l,S_{1},...,S_{N})=\prod_{n=1}^{N}p(S_{n}|l,S_{1},...,S_{n-1}).
\end{equation}

Second, the spatial autoregressive model generates each frame in a set-by-set order.
The generation of token sets within the n-th frame is formulated as:
\begin{equation}
\begin{gathered}
p(S_{n}^{\prime},S_{(n,1)},...,S_{(n,K)})\\=\prod_{k=1}^{K}p(S_{(n,k)}|S_{n}^{\prime},S_{(n,1)},...,S_{(n,k-1)}),
\end{gathered}
\end{equation}
where \(S_{(n,k)}\) denotes the k-th token set of n-th frame.

Finally, the diffusion loss as in~\cref{eq:diffusion_loss} is applied to estimate per-token probability in a continuous space, with $z_{(n,k)}=S_{(n,k)}$.
\section{Method}
\label{sec:cmast_method}
\subsection{Overview}

PhysGen is an autoregressive framework that operates on a variable-length history of visual observations and actions.
The overall architecture is shown in~\cref{fig:ar_transformer}.
At each step, the model takes as input a task instruction $l$, a sequence of past images $ \left\{ O_0, O_1, O_2, \cdots, O_{N-1} \right\} $ and corresponding action chunks $ \left\{ A_1, A_2, \cdots, A_{N-1} \right\} $. It outputs the next visual state $ O_N $ together with the corresponding action chunk $ A_N $ for execution.
An action chunk $A_n$ consists of $L$ consecutive actions, as in prior works~\cite{chi2023diffusion, zhao2023learning}.
This autoregressive formulation allows the model to function as a predictive world interaction model, explicitly capturing the step-by-step evolution of visual observations and actions.

\subsection{Model Architecture}
Our PhysGen model is built upon the video autoregressive generation model, NOVA\cite{deng2024autoregressive}, transferring pretrained world knowledge from NOVA’s representations into joint video-action physics evolution.
Specifically, PhysGen tokenizes vision and actions into physical tokens, models them autoregressively, and de-tokenizes them into predicted video and actions.
This design supports a unified perception–planning–execution loop within continuous space.

\subsubsection*{Tokenizer}

To fully preserve NOVA’s pretrained knowledge, PhysGen adheres to its original embedding schemes for language and vision.
Specifically, task instructions are tokenized by a frozen Phi language model~\cite{javaheripi2023phi} to generate language token $E_l \in \mathbb{R}^{K_l \times d}$, while visual observations are tokenized by a frozen 3D-VAE~\cite{zheng2024open} and flattened into frame tokens $E_{O,n} \in \mathbb{R}^{K_O \times d}$.
In addition, an action tokenizer is introduced to project action chunks into the unified continuous physical embedding space via an MLP, yielding  $K_A = L$ action tokens $E_{A,n} \in \mathbb{R}^{K_A \times d}$ that facilitate joint processing by the physical autoregressive model.

\subsubsection*{Physical Autoregression}

We adopt the standard token-by-token autoregressive paradigm used in large language models.
The difference is that PhysGen operates over \emph{physical tokens} that jointly represent visual states and actions, enabling unified step-by-step modeling of the perception–prediction–action loop.
Specifically, each physical token is constructed by concatenating the respective frame and action tokens along the sequence dimension, resulting in the joint representation:
\begin{equation}
P_{n} = \left[ E_{O,n}; E_{A,n}\right]; \quad P_{n} \in \mathbb{R}^{(K_O + K_A) \times d}.
\end{equation}
To account for the one-step temporal offset where observations lead actions, a learnable Begin of Action (BOA) token is prepended to the action sequence to align their lengths.
The autoregression process then predicts the next physical token conditioned on the sequence of preceding tokens, formulated as:
\begin{equation}
    p\left( E_l,P_0,\cdots ,P_N \right)=\prod_{n=0}^N{p\left( P_n|E_l,P_0,\cdots ,P_{n-1} \right)}.
\label{eq:sequence_distribution}
\end{equation}
where the conditional distribution is parameterized by a Causal Transformer network that replicates the architecture of NOVA.

\subsubsection*{De-Tokenizer}

A central challenge in de-tokenization is estimating the conditional distribution of output tokens.
Prior works typically map discrete vocabularies to images or actions~\cite{zitkovich2023rt, zheng2024open, wang2025unified}, enabling likelihood computation via categorical distributions but introduces quantization errors~\cite{zhang2025chain}.
Alternatively, some approaches project continuous tokens directly by linear layers or MLPs~\cite{wu2023unleashing, fu2024context, zhang2025chain}, avoiding quantization but sacrificing generative sampling.
In this work, akin to MAR~\cite{li2024autoregressive}, we estimate conditional distributions through diffusion process, maintaining a representation that is both continuous and generative.

Specifically, the Transformer outputs are taken as the conditional vector $Z_n$, encapsulating the entire preceding context:
\begin{equation}
Z_n = Transformer(l, P_0, \cdots, P_{n-1}).
\end{equation}
Then, a DiT-based denoising process is performed, conditioned on $Z_n$, to estimate the conditional distribution $p(P_n | Z_n)$, formulated as:
\begin{equation}
\mathcal{L}(P_n,Z_n)=\mathbb{E}_{\epsilon,t}[||\epsilon-\epsilon_{\theta}(P_{n,t}|t,Z_n)||^{2}].
\label{eq:diffusion_loss_ar}
\end{equation}
The gradient is backpropagated to $Z$ for optimizing the parameters in the Transformer backbone.

In the inference stage, token $P_n = P_{n,t=0}$ is sampled through the reverse diffusion process guided by the corresponding $Z_n$:
\begin{equation}
P_{n,t-1}=\frac{1}{\sqrt{\alpha_{t}}}\left(P_{n,t}-\frac{1-\alpha_{t}}{\sqrt{1-\overline{\alpha}_{t}}}\epsilon_{\theta}(P_{n,t}|t,Z_n)\right)+\sigma_{t}\delta.
\end{equation}
The denoised tokens $P_{n,t=0}$ are subsequently transformed into the predicted images and actions.

The de-tokenization of frame and action tokens is performed separately.
For frame tokens, we adopt the reconstruction paradigm from NOVA.
For action tokens, we introduce Action-DiT, a lightweight DiT network for denoising, with conditioning vectors injected through cross-attention.

\begin{figure}[t]
    \centering
    \includegraphics[width=0.96\linewidth]{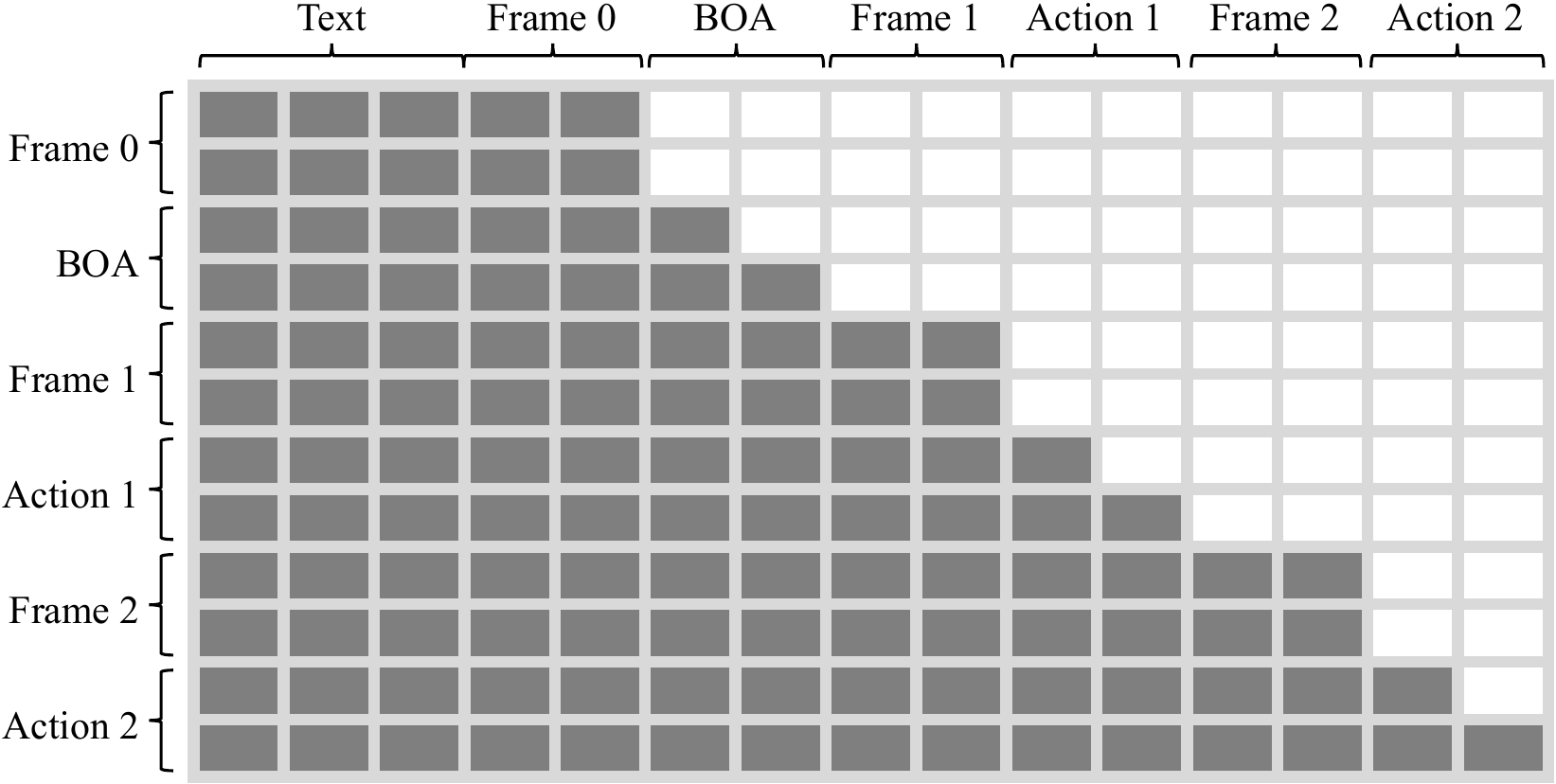}
    \caption{\underline{The causal attention mask.} Frame tokens use chunk-wise full attention; action tokens use temporal causal attention; action tokens unidirectionally attend to frame tokens for visually conditioned control.}
    \label{fig:causal_mask}
\end{figure}

\begin{table*}[t]
\centering
\caption{\underline{The success rate comparison} in the Libero Benchmark. The best are highlighted by \textbf{bold}. `No action pretraining' means that the model’s backbone is not pretrained on any large-scale manipulation datasets.}
\resizebox{0.83\textwidth}{!}{
\begin{tabular}{@{}lcccccc@{}}
\toprule
Method                     & \multicolumn{1}{c|}{Parameters}                      & Spatial & Object & Goal   & \multicolumn{1}{c|}{Long}                           & Average    \\ \midrule
\rowcolor[HTML]{F2F2F2} 
\multicolumn{7}{l}{\cellcolor[HTML]{F2F2F2}\textit{\qquad \qquad Training from Scratch Baseline}}                                                                                           \\
DP \citeyearpar{chi2023diffusion} (w/o act. pretraining)  & \multicolumn{1}{c|}{--}                         & 78\%    & 93\%   & 68\%   & \multicolumn{1}{c|}{51\%}                           & 72\%   \\
\rowcolor[HTML]{F2F2F2} 
\multicolumn{7}{l}{\cellcolor[HTML]{F2F2F2}\textit{\qquad \qquad Vision-Language-Pretrained Baseline}}                                                                                 \\
Octo \citeyearpar{team2024octo}                       & \multicolumn{1}{c|}{93M}                         & 79\%    & 86\%   & 85\%   & \multicolumn{1}{c|}{51\%}                           & 75\%   \\
TraceVLA \citeyearpar{zheng2024tracevla}                   & \multicolumn{1}{c|}{7B}                         & 85\%    & 85\%   & 75\%   & \multicolumn{1}{c|}{54}                           & 75\%   \\
OpenVLA \citeyearpar{kim2024openvla}                   & \multicolumn{1}{c|}{7B}                         & 85\%    & 88\%   & 79\%   & \multicolumn{1}{c|}{54\%}                           & 77\%   \\
SpatialVLA \citeyearpar{qu2025spatialvla}                & \multicolumn{1}{c|}{4B}                         & 88\%    & 90\%   & 79\%   & \multicolumn{1}{c|}{56\%}                           & 78\%   \\
ThinkAct \citeyearpar{huang2025thinkact}                  & \multicolumn{1}{c|}{7B}                         & 88\%    & 91\%   & 87\%   & \multicolumn{1}{c|}{71\%}                           & 84\%   \\
Pi0-Fast \citeyearpar{pertsch2025fast}                  & \multicolumn{1}{c|}{3B}                         & \textbf{96}\%    & 97\%   & 89\%   & \multicolumn{1}{c|}{60\%}                           & 86\%   \\
MolmoACT \citeyearpar{lee2025molmoact}                  & \multicolumn{1}{c|}{7B}                         & 87\%    & 95\%   & 88\%   & \multicolumn{1}{c|}{77\%}                           & 87\%   \\
\rowcolor[HTML]{F2F2F2} 
\multicolumn{7}{l}{\cellcolor[HTML]{F2F2F2}\textit{\qquad \qquad Visual-Generation-Pretrained Baseline}}                                                                                \\
UniMimic \citeyearpar{chen2025unifying} (w/o act. pretraining) \qquad \qquad & \multicolumn{1}{c|}{$\sim$200M}                         & 71\%    & 79\%   & 67\%   & \multicolumn{1}{c|}{29\%}                           & 62\%   \\
UniMimic \citeyearpar{chen2025unifying}                   & \multicolumn{1}{c|}{$\sim$200M}                         & 89\%    & 91\%   & 85\%   & \multicolumn{1}{c|}{59\%}                           & 81\%   \\
CoT-VLA \citeyearpar{zhao2025cot}                   & \multicolumn{1}{c|}{7B}                         & 88\%    & 92\%   & 88\%   & \multicolumn{1}{c|}{69\%}                           & 84\%   \\
WorldVLA \citeyearpar{cen2025worldvla} (w/o act. pretraining) & \multicolumn{1}{c|}{7B}                         & 88\%    & 96\%   & 83\%   & \multicolumn{1}{c|}{60\%}                           & 82\%   \\ \midrule
\rowcolor[HTML]{E8F5E9} 
PhysGen (Ours, without action pretraining) & \multicolumn{1}{c|}{\cellcolor[HTML]{E8F5E9}732M} & 91.0\%  & \textbf{99.6}\% & \textbf{93.8}\% & \multicolumn{1}{c|}{\cellcolor[HTML]{E8F5E9}\textbf{78.8}\%} & \textbf{90.8}\% \\ \bottomrule
\end{tabular}
}
\label{tab:comparison_libero}
\end{table*}

\subsection{Model Design Choices}
Beyond the core autoregressive architecture, we introduce several design choices that facilitate joint autoregressive modeling.

\subsubsection*{Causal Mask}
Our causal masking scheme is tailored to the joint modeling of frames and actions, as illustrated in~\cref{fig:causal_mask}.
Each physical token consists of a frame part and an action part.
For the frame part, we employ a chunk-based attention mask, which allows all patches within the same frame to attend to each other.
For the action part, we use a temporal causal mask so that earlier actions within a chunk cannot attend to later actions.
Additionally, action tokens are allowed to attend to frame tokens in a unidirectional manner, which implicitly captures inverse kinematics by conditioning action planning on future visual states.
Across chunks, a temporal causal mask is enforced to maintain temporal causality.

\subsubsection*{Lookahead-MTP}
Prior works have shown that lookahead formulations enable planning over a longer temporal window, resulting in more coherent action sequences
~\cite{chi2023diffusion, black2024pi0, zhang2025dreamvla, kim2025fine}.
Inspired by this, we introduce Lookahead Multi-Token Prediction (L-MTP), which integrates lookahead action generation into the Multi-Token Prediction paradigm~\cite{gloeckle2024better, liu2024deepseek, yang2025qwen3}.
Specifically, at each autoregressive step, the action de-tokenizer generates multiple future tokens in parallel (implemented with 3 tokens).
During training, all predicted tokens are supervised.
At inference time, only the first predicted token is executed, while the remaining tokens serve as lookahead information to condition subsequent predictions.

\subsubsection*{Efficient Training and Inference}
For training, we adopt a fully parallelized strategy with teacher forcing, computing losses for all tokens in a single forward pass.
To further improve training efficiency, we fine-tune the Transformer backbone using LoRA.
During inference, we employ a KV-cache mechanism~\cite{NVIDIA2023LLM} to cache intermediate features at each layer, enabling efficient autoregressive generation.

\subsection{Loss Function}
The training objective is defined as the average diffusion loss over the sequence of physical tokens:
\begin{equation}
\begin{aligned}
&loss=\sum_{n=1}^N\mathcal{L}(Z_n,P_n)\\&=\sum_{n=1}^N\left(\mathcal{L}_{obs}(Z_{n},E_{O,n}) + \mathcal{L}_{act}(Z_{n},E_{A,n})\right).
\end{aligned}
\end{equation}

\begin{table}[t]
    \centering
    \caption{\underline{The success rate comparison} in the ManiSkill Benchmark. The best are highlighted by \textbf{bold}.}
    \setlength{\tabcolsep}{2pt}
    \begin{tabular}{lcccc}
      \toprule
      Method \qquad \qquad \qquad & PushCube & PickCube & StackCube & Avg. \\
      \midrule
ACT [\citeyear{zhao2023learning}] & 76\% & 20\% & 30\% & 42\% \\
BC-T [\citeyear{mandlekar2021matters}] & 98\% & 4\% & 14\% & 39\% \\
DP [\citeyear{chi2023diffusion}] & 88\% & 40\% & \textbf{80}\% & 69\% \\
ICRT [\citeyear{vuong2025action}] & 77\% & \textbf{78}\% & 30\% & 62\% \\
RDT [\citeyear{liurdt}]  & \textbf{100}\% & 77\% & 74\% & \textbf{84}\% \\
Pi0 [\citeyear{black2024pi0}]  & \textbf{100}\% & 60\% & 48\% & 69\% \\
      \midrule
\rowcolor[HTML]{E8F5E9}
PhysGen (Ours) & \textbf{100}\% & 73\% & 48\% & 74\% \\
      \bottomrule
    \end{tabular}
    \label{tab:comparison_maniskill}
\end{table}

\subsection{Implementation Details}

The chunk size of each action chunk is set to 8 (i.e. $L = 8$).
The maximum context length of our Transformer Backbone is set to 2096.
Specifically, it includes 256 language tokens and 5 physical token packages, where each physical package contains 360 visual tokens and 8 action tokens.
Multi-view observations are concatenated into a single image and fed directly into the VAE and the autoregressive transformer, allowing the model’s inherent self-attention to maintain cross-view consistency.

PhysGen’s backbone is not pretrained on any large-scale action or manipulation datasets (what we term \emph{\textbf{no action pretraining}}); it is pretrained only on video generation and then directly fine-tuned on downstream manipulation tasks.
All finetuning experiments are conducted on a single NVIDIA A100-SXM4-80GB GPU, with the longest training completed within 60 GPU hours.
During training, we follow~\cite{li2024autoregressive} to sample $t$ by 4 times for each image and each action.
We employ the Rotary Positional Embeddings (RoPE)~\cite{su2024roformer} to preserve the temporal dependencies, with distinct frequency settings for frame and action tokens.

\section{Experiment}
\label{sec:experiment}

\subsection{Simulation Experiment}
We conduct simulation experiments on the LIBERO~\cite{liu2023libero} and ManiSkill~\cite{taomaniskill3} benchmarks, using success rate as the primary metric.
For LIBERO, we evaluate four task suites: \emph{LIBERO-Spatial}, \emph{LIBERO-Object}, \emph{LIBERO-Goal}, and \emph{LIBERO-Long}.
Following established protocols~\cite{zheng2024open, cen2025worldvla}, each suite is fine-tuned using approximately 400 demonstrations and evaluated over 500 distinct rollouts.
For ManiSkill, we focus on three representative tasks: \emph{PushCube}, \emph{PickCube}, and \emph{StackCube}.
We follow the corresponding protocols~\cite{liurdt, vuong2025action}, using 1000 demonstrations per task for fine-tuning and 125 distinct rollouts for evaluation.

We compare our method against a comprehensive set of SOTA baselines, including: (1) \emph{Vision-Language-Pretrained Baselines}, which are built upon pretrained LLMs/VLMs, such as OpenVLA~\cite{zheng2024open}; (2) \emph{Visual-Generation-Pretrained Baselines}, which are built upon pretrained visual generation models, such as WorldVLA~\cite{cen2025worldvla}; and (3) other representative architectures such as RDT~\cite{liurdt}.
Crucially, while some baselines are pretrained on extensive action demonstrations, our method and those marked use backbones that are not pretrained on any large-scale manipulation datasets (i.e., no action pretraining), highlighting our efficiency.
All reported metrics are derived from official papers or validated reproductions~\cite{lee2025molmoact, wang2025vla, zhan2025mathcal}.
The results are summarized in Table~\ref{tab:comparison_libero} and Table~\ref{tab:comparison_maniskill}.

\subsubsection*{Analysis on LIBERO}
On LIBERO, PhysGen achieves the highest average success rate among all compared methods, outperforming both WorldVLA~\cite{cen2025worldvla} and Pi0-Fast~\cite{pertsch2025fast}.
Compared with WorldVLA~\cite{cen2025worldvla}, PhysGen achieves an average gain of 8.8 percentage points, with an 18.8-point improvement on the long-horizon LIBERO-Long tasks.
Although both methods adopt a joint image–action generation paradigm, WorldVLA relies on discrete tokenization, whereas PhysGen models both visual and action in a shared continuous embedding space.
Similarly, CoT-VLA~\cite{zhao2025cot} also adopts discrete visual tokenization and falls behind PhysGen by 9.2\% in absolute success rate.
Given that images and actions are inherently continuous signals, our continuous formulation better preserves their structure and promotes tighter cross-modal interaction, enabling more seamless transfer of physical knowledge from video generation to action prediction.
Compared with Pi0-Fast~\cite{pertsch2025fast}, PhysGen surpasses it by an absolute margin of 4.8\% in average success rate.
While Pi0-Fast relies on a pretrained vision–language backbone, PhysGen leverages a pretrained video generation backbone, achieving superior performance with enhanced data efficiency.
This result demonstrates that video-generative priors are particularly effective for robotic manipulation and suggesting a promising direction for exploiting such models.
Finally, PhysGen underperforms Pi0-Fast only on LIBERO-Spatial, likely due to the limited spatial perception of the underlying video model, which we identify as an avenue for future improvement.

\begin{table}[]
\caption{\underline{Real world performance}. We report the task success rates on four tabletop tasks. PhysGen achieves an average success rate comparable to Pi0.}
\begin{tabular}{@{}lccccc@{}}
\toprule
Method     & \begin{tabular}[c]{@{}c@{}}Pick\\ Cube\end{tabular} & \begin{tabular}[c]{@{}c@{}}Press\\ Button\end{tabular} & \begin{tabular}[c]{@{}c@{}}Stack\\ Cube\end{tabular} & \begin{tabular}[c]{@{}c@{}}Pick\\ Transparency\end{tabular} & Avg   \\ \midrule
ACT \cite{zhao2023learning}        & 40\%        & 40\%           & 30\%         & 10\%                & 30\%    \\
OpenVLA \cite{kim2024openvla}    & 30\%        & 25\%           & 10\%         & 0\%                 & 16.3\% \\
Pi0 \cite{black2024pi0}        & \textbf{85}\%        & \textbf{85}\%           & \textbf{60}\%         & 70\%                & \textbf{75}\%    \\
PhysGen (Ours) & 80\%        & \textbf{85}\%           & \textbf{60}\%         & \textbf{75}\%                & \textbf{75}\%    \\ \bottomrule
\end{tabular}
\label{tab:real_world_performance}
\end{table}

\begin{table}[]
\caption{\underline{The ablation study} on the Libero-Object task of the Libero Benchmark. SR denotes Success Rate; higher is better.}
\resizebox{0.48\textwidth}{!}{
\begin{tabular}{@{}lcccc|c@{}}
\toprule
Method       & Pretrain & Token & L-MTP & Backbone & SR \\ \midrule
PhysGen-Zero     & no       & cont.            & +     & AR                    & 86.4\%    \\
PhysGen-Discrete & NOVA     & discrete         & +     & AR                    & 94.2\%    \\
PhysGen-NoAR     & NOVA     & cont.            & -     & NoAR                  & 95.0\%    \\
PhysGen-STP      & NOVA     & cont.            & -     & AR                    & 96.8\%    \\
PhysGen-Full     & NOVA     & cont.            & +     & AR                    & 99.6\%    \\ \bottomrule
\end{tabular}
}
\label{tab:ablation_study}
\end{table}

\subsubsection*{Analysis on ManiSkill}
On ManiSkill, PhysGen achieves an average success rate comparable to or exceeding several methods that rely on large-scale action pretraining. Specifically, while slightly trails RDT~\cite{liurdt}, PhysGen outperforms ICRT~\cite{vuong2025action} and Pi0~\cite{black2024pi0} by absolute margins of 12\% and 5\%, respectively.
Notably, PhysGen achieves a perfect 100\% success rate on the PushCube task, underscoring its robustness.
These results highlight the effectiveness of transferring physical priors from pretrained video generation models. It suggests that sophisticated manipulation competence can be acquired without exhaustive action-labeled datasets, significantly enhancing data efficiency for downstream control.

\begin{figure*}[t]
    \centering
    \includegraphics[width=0.95\linewidth]{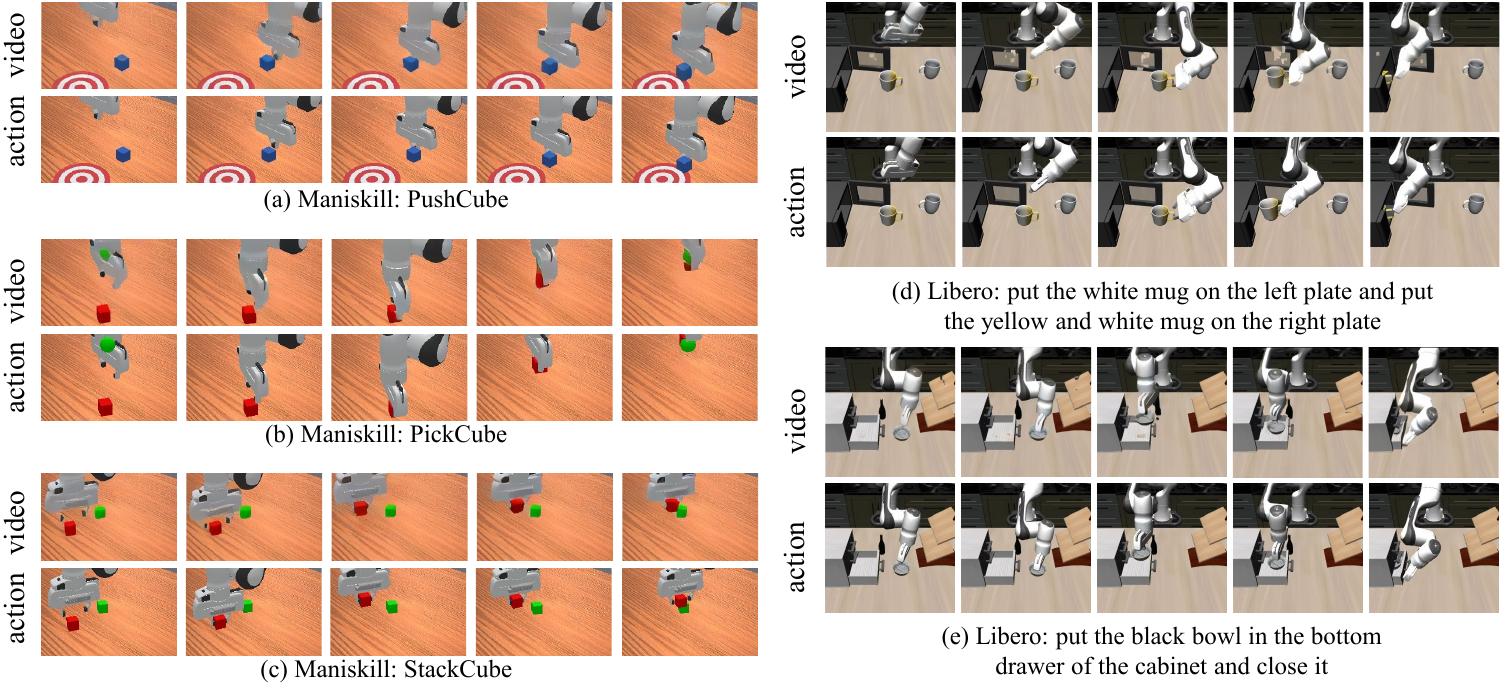}
    \caption{\underline{Video predictions and actual action executions.} Each row shows PhysGen's predicted video alongside the corresponding execution video for five different tasks. The strong visual similarity between predicted and actual action videos highlights the effectiveness of our approach in transferring knowledge from video pretraining.}
    \label{fig:vis_success}
\end{figure*}

\begin{figure*}[]
    \centering
    \includegraphics[width=0.9\linewidth]{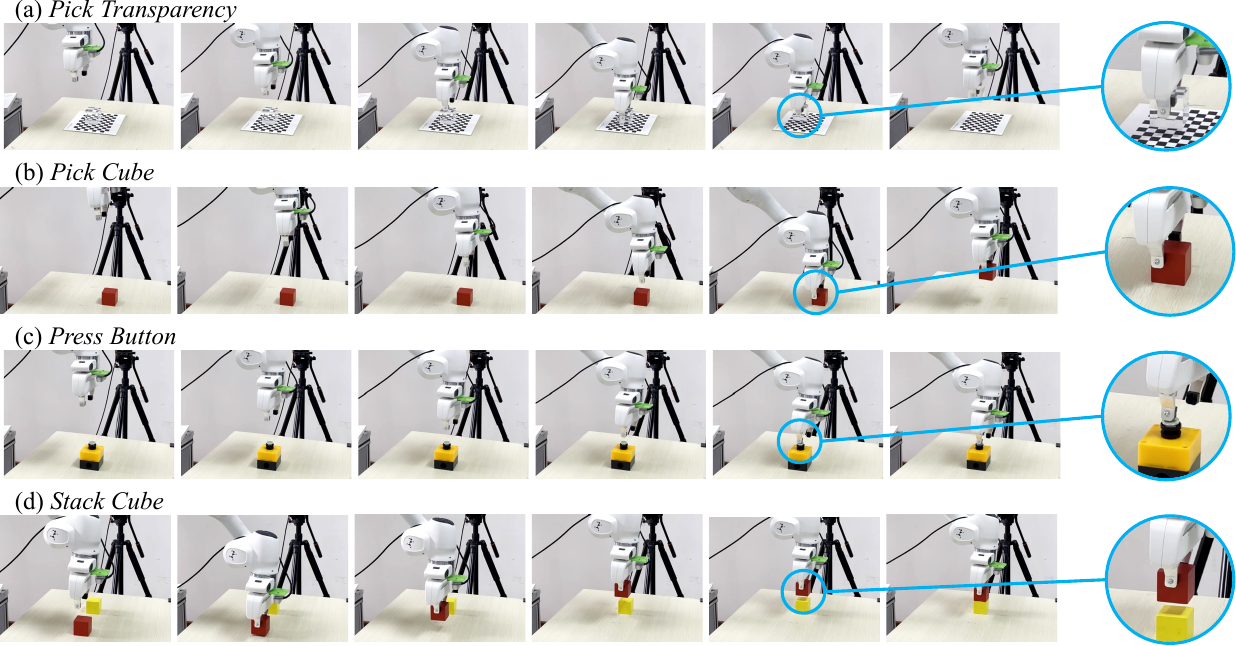}
    \caption{\underline{Visualization of real world manipulation.} We deploy our method on a Franka Panda robot across four real-world tasks: \emph{Pick Transparency}, \emph{Pick Cube}, \emph{Press Button}, and \emph{Stack Cube}. The robot executes stable and coherent manipulations across diverse settings, demonstrating effective knowledge transfer from video generation to real-world manipulation.}
    \label{fig:vis_realworld}
\end{figure*}

\subsection{Real World Experiment}
We conduct real-world experiments using a Franka Panda arm with two RealSense D415 cameras, one fixed and one wrist-mounted.
Following previous practices~\cite{xu2025a0, zhang2025dreamvla}, we evaluate on four tasks: \emph{Pick Cube}, \emph{Press Button}, \emph{Stack Cube}, \emph{Pick Transparency} (see Appendix for task details). The \emph{Pick Transparency} evaluates physical awareness by requiring grasping a transparent cube.

We collect 80 to 100 teleoperated demonstrations per task for training and evaluate 20 trials per task with different initializations for evaluation.
Our method is trained solely on these collected data without pre-training.
For comparison, we benchmark against ACT~\cite{zhao2023learning}, OpenVLA~\cite{zheng2024open} and Pi0~\cite{black2024pi0}.
ACT is trained from scratch, OpenVLA and Pi0 are finetuned from their officially pretrained checkpoints.
The quantitative results are reported in~\cref{tab:real_world_performance}, the rollouts are shown in~\cref{fig:vis_realworld}.

As shown in~\cref{tab:real_world_performance}, PhysGen achieves an average success rate comparable to Pi0~\cite{black2024pi0} and surpasses OpenVLA~\cite{kim2024openvla} across four real-world tasks, even though its backbone is not pretrained on any large-scale manipulation datasets.
In contrast, both Pi0 and OpenVLA rely on large-scale action-pretrained backbones, whereas PhysGen leverages a pretrained autoregressive video generation backbone and transfers its physical knowledge to downstream manipulation.
These results indicate that such video-derived knowledge provides a strong prior for real-world control, enabling competitive performance with substantially improved data efficiency.
In particular, PhysGen outperforms Pi0 on the \emph{Pick Transparency} task, which requires grasping a transparent cube.
Such conditions demand stronger physical knowledge, since refraction and reflection create ambiguous visual observations.
PhysGen achieves a 5 percentage point higher success rate than Pi0.
This improvement suggests that physical intuition acquired from video generation pretraining helps PhysGen better handle physically demanding manipulation.

\subsection{Ablation Study}
We conduct ablation studies on the Libero-Object task the Libero Benchmark~\cite{liu2023libero} to isolate the contributions of our design choices.
The results are summarized in~\cref{tab:ablation_study}.

\subsubsection*{Video Generation Pretraining}
To investigate the role of video generation pretraining, we introduce an ablated variant, PhysGen-Zero, which uses the same architecture as PhysGen-Full without loading pretrained weights.
As shown in our experiments, PhysGen-full outperforms PhysGen-zero by a 13.2\% absolute margin in success rate, highlighting the critical role of video pretraining.
This result indicates that dynamic priors learned from video generation pretraining enable PhysGen to perform physically grounded manipulation.

\subsubsection*{Token Representation}

To evaluate the benefit of joint continuous joint representation, we introduce PhysGen-Discrete, which quantizes action signals into a fixed vocabulary as in OpenVLA~\cite{zheng2024open} and WorldVLA~\cite{cen2025worldvla}, while retaining continuous visual tokens to adhere to the NOVA setting.
As shown in our experiments, PhysGen-Full outperforms PhysGen-Discrete by a 5.4\% absolute margin in success rate.
This result suggests that quantization introduces resolution errors that hinder action prediction, while a shared continuous embedding space facilitates effective interaction and knowledge transferring between vision and action, leading to more precise manipulation.

\subsubsection*{Autoregressive Architecture}
We introduce the PhysGen-NoAR (NoAR denotes No-Autoregressive) to examine the role of the autoregressive architecture.
PhysGen-NoAR removes the autoregressive rollout and maps a single-step input to a single-step output in an end-to-end manner, which is equivalent to setting the context horizon $N=1$ in~\cref{eq:sequence_distribution}.
PhysGen-Full outperforms PhysGen-NoAR by 4.6\% absolute margin in success rate.
This result highlights the importance of the autoregressive learning paradigm, which captures the iterative evolution of both the environment and the embodiment over time, producing temporally coherent planning and smoother action trajectories.

\subsubsection*{Lookahead-MTP}
We remove the L-MTP module to assess its contribution, yielding a Single-Token Prediction variant, denoted as PhysGen-STP.
As shown in our experiments, eliminating L-MTP results in a 3.4 percentage point drop in absolute success rate.
We hypothesize two underlying reasons for this performance drop.
(1) Multi-token prediction promotes more stable training convergence, as demonstrated in LLMs~\cite{liu2024deepseek}.
(2) The lookahead formulation provides a longer planning horizon, which improves task success.


\begin{figure}[t]
    \centering
    \includegraphics[width=0.9\linewidth]{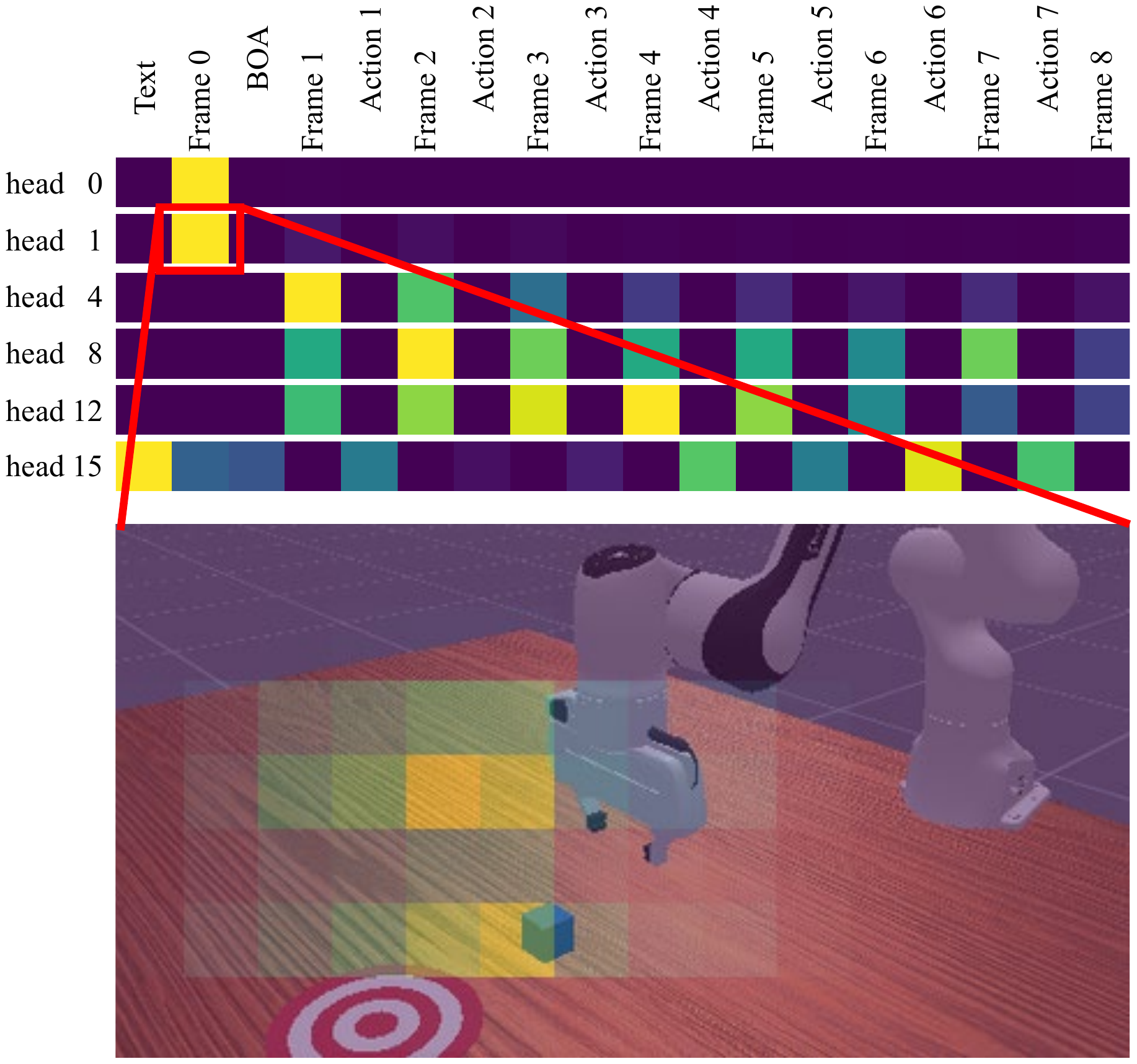}
    \caption{\underline{The attention map.} The top row shows the token-level attention map, indicating how the predicted action attends to previous frame and action tokens. The bottom row shows the pixel-level attention map, indicating how the predicted action attends to different spatial regions of the frame.}
    \label{fig:attn_map}
\end{figure}

\subsection{Qualitative Analysis}

\subsubsection*{Video Action Alignment}
We visualize the predicted video sequences alongside the corresponding execution videos in~\cref{fig:vis_success}. The results reveal a clear alignment between the predicted visual trajectories and actual robot actions, with highly similar motion trajectories and timing of key actions.
For example, in the second row of the PickCube task, the predicted video correctly captures the rotation of the arm to match the cube’s orientation, a meticulous step for a successful grasping.
The actual execution precisely mirrors this behavior, rotating by the correct angle to match the cube’s orientation and enable a successful grasping.
This consistency highlights not only the model’s fine-grained scene understanding from video pretraining, but also its effectiveness in transferring that understanding into action planning.

\subsubsection*{Real World Manipulation}
We provide qualitative visualizations of PhysGen’s execution on real-world tasks in \cref{fig:vis_realworld}. The results show stable and coherent manipulation behaviors across diverse scenarios.
Notably, PhysGen reliably completes the \emph{Pick Transparency} task despite the complex and misleading visual patterns produced by refraction and reflection in the transparent crystal cube.
The robot accurately localizes and grasps the crystal cube, demonstrating robust behavior under visually challenging conditions.
These results further demonstrate that physical intuition acquired through video generation can be effectively transferred to real-world manipulation.

\subsubsection*{Attention Map Visualization}
We visualize the attention maps of PhysGen to better understand how it attends to visual and action cues during prediction, as shown in~\cref{fig:attn_map}. At the token level, different attention heads selectively focus on frame tokens or action tokens, indicating that the model effectively captures the interplay between perception and action for joint modeling. At the pixel level, the attention is concentrated on task-critical regions, such as the cube, target area, and the robotic arm, demonstrating that PhysGen can attend to spatially relevant features that directly influence task execution.
\section{Conclusion}
\label{sec:conclusion}

This work introduces PhysGen, a physical autoregressive framework that repurposes pretrained video generation models as predictive world interaction models for robotic manipulation.
By jointly modeling the coupled evolution of visual observations and actions through shared physical tokens, PhysGen transfers implicit physical knowledge from video pretraining to action generation, enabling physically grounded control.
A multimodal continuous representation is employed to further unify frames and actions in a shared space, promoting fine-grained modeling and effective cross-modal interaction.
Additionally, we incorporate inverse-kinematics causal mask, Lookahead Multi-Token Prediction, along with efficient training and inference strategies.
Across extensive evaluations on ManiSkill and LIBERO, PhysGen consistently outperforms strong baselines, surpassing OpenVLA and WorldVLA by 13.8\% and 8.8\% in absolute success rate, respectively.
Furthermore, it demonstrates stable and robust manipulation in real-world settings, even for physically challenging tasks, while maintaining high data efficiency.
Qualitative results reveal tightly aligned video and action predictions, reflecting an accurate modeling of world–robot dynamics.
Overall, PhysGen underscores the potential of repurposing autoregressive video models as sequential world interaction models, offering a promising direction for developing generalizable robotic intelligence.


\clearpage
\bibliographystyle{ACM-Reference-Format}
\bibliography{main}

\clearpage
\appendix

\end{document}